\documentclass[sigconf,nonacm]{acmart}

\setcopyright{none}
\acmConference{}{}{}
\acmBooktitle{}
\acmDOI{}
\acmISBN{}
\usepackage{amsmath}
\usepackage{booktabs}
\usepackage{graphicx}
\usepackage{microtype}
\usepackage{placeins}
\usepackage{balance}

\newtheorem{proposition}{Proposition}
\newtheorem{corollary}{Corollary}

\newcommand{\E}{\mathbb{E}}
\newcommand{\Cov}{\operatorname{Cov}}
\newcommand{\Var}{\operatorname{Var}}

\title{The Retry Tax: Why Static Same-Task Retries Can Cost More per Completion}

\author{Wei-Jung Huang}
\affiliation{%
  \institution{Independent Researcher}
  \city{New York}
  \state{NY}
  \country{USA}
}
\email{william.wj.huang@gmail.com}

\begin{document}
\raggedbottom

\begin{abstract}
Retrying an agent after failure can complete more tasks while increasing the cost per completion. We study a static policy that stops at verified success or an attempt cap and otherwise resets the original prompt and workspace without passing feedback to the next request. Under stationary task success probabilities, independent attempts conditional on the task, and a common attempt cost, easier tasks tend to succeed earlier. When task success probabilities differ, the unresolved mix shifts toward harder tasks. Specializing classical survivor selection, we show that cost per completion is at least as high as in one-shot serving and that marginal cost per additional completion is nondecreasing wherever the completion increments are positive. These are different quantities: one averages spending over all completions, while the other prices the completions added by another attempt. We then arrange each task's existing benchmark runs in different orders, stop each sequence at verified success or the attempt cap, and average the resulting completion and spending at each supported cap. For one OpenHands configuration, replay estimates completion of 59.7 percent with one attempt and 70.0 percent with four, while cost per completion rises from about \$0.70 to \$1.32. The recorded trade-off describes fixed configurations on the available tasks. Choosing a cap requires comparing the additional spending with the application's value for the additional completions.
\end{abstract}



\maketitle
\pagestyle{plain}
\thispagestyle{plain}

\section{Introduction}

A code-agent service must decide whether to try again after a failed patch. Another attempt may complete the task, but the service also pays for attempts that fail. We study a \emph{static} same-task retry policy: restore the original inputs after failure, and stop at verified success or a fixed cap. A cap of four permits at most four attempts in total, including the first. The decision is how much to spend for the additional completions.

\begin{figure*}[t]
  \centering
  \includegraphics[width=\textwidth]{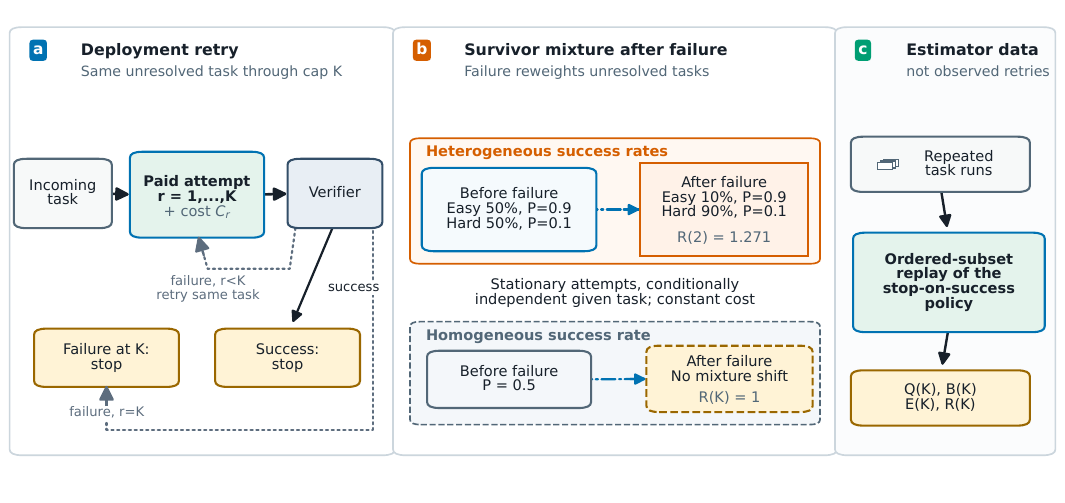}
  \caption{Static retries and their measurement. (a) Success stops the service; failure permits another attempt on the same task through cap $K$. (b) The equal-weight easy/hard mixture and the homogeneous population have the same one-shot success rate. Displayed mixtures and ratios use stationary success probabilities, conditional independence, and constant attempt cost. (c) Averaging stopped orders of recorded outcome-cost pairs gives completion $Q$, stopped spend $B$, cost per completion $E$, and the ratio $R$ to one-shot serving.}
  \Description{Three panels show the stop-on-success retry rule, the changing unresolved task mix in a heterogeneous example alongside a homogeneous control, and the calculation from repeated recorded runs.}
  \label{fig:method}
\end{figure*}

When task success probabilities differ, later attempts can face a different mix even when the agent and the price per attempt stay fixed. Suppose half the tasks have success probability 0.9 and half have probability 0.1, with independent attempts conditional on the task. After one failure, the expected mix is 10 percent easier tasks and 90 percent harder tasks. Easier tasks tend to succeed earlier, leaving a larger share of harder tasks eligible for another attempt (Figure~\ref{fig:method}). Section~\ref{sec:selection} works through the spending consequences.

Repeated benchmark runs make the cost comparison concrete. Table~\ref{tab:intro-example} gives replay estimates for one fixed OpenHands configuration on 500 SWE-bench Verified tasks~\cite{bai2026agentspend}. The larger cap completes more tasks at a higher average cost per completion.

\begin{table}[t]
\centering
\caption{Recorded-run replay for one OpenHands configuration on 500 tasks.}
\label{tab:intro-example}
\begin{tabular}{lrr}
\toprule
Quantity & One attempt & Up to four attempts \\
\midrule
Completion & 59.7\% & 70.0\% \\
Cost per completion & \$0.70 & \$1.32 \\
\bottomrule
\end{tabular}
\end{table}

The model specializes classical survivor selection in heterogeneous populations~\cite{vaupel1985ruses,proschan1963dfr,alvarez2024duration}. The paper makes two contributions:
\begin{itemize}
  \item Under stationary task probabilities, conditional independence, and common cost, we derive covariance and cap-two variance identities for cost per completion. We characterize marginal cost and when adjacent comparisons identify the best common cap.
  \item For three panels of public repeated runs, we calculate completion and stopped spend at every supported cap for each fixed configuration. These recorded comparisons separate average cost per completion from the cost of additional completions.
\end{itemize}

\section{Why Later Completions Can Cost More}
\label{sec:theory}

\subsection{Completion, spend, and average cost}
\label{sec:controller}

Fix an agent configuration, environment, verifier, and incoming task distribution. After each attempt, the controller observes the verifier's outcome and stops at success or the cap. If another attempt is allowed, the actor receives the original prompt and reset workspace, with no information from prior runs. Resetting specifies the actor's inputs; stationarity and conditional independence remain additional assumptions about its outcomes.

Let $S_r\in\{0,1\}$ be the verifier outcome and $C_r>0$ the cost of attempt $r$. The policy quantities are
\begin{align}
Q(K) &= \Pr\!\left(\max_{1\le r\le K} S_r=1\right), \\
B(K) &= \E\!\left[\sum_{r=1}^{K} C_r\prod_{s<r}(1-S_s)\right], \\
E(K) &= \frac{B(K)}{Q(K)}, \qquad R(K)=\frac{E(K)}{E(1)}.
\end{align}
The product in $B(K)$ charges attempt $r$ only when every earlier attempt failed. Thus $B(K)$ includes the spend on tasks still unresolved at the cap, while $E(K)$ divides all stopped spend by the completed share. A value $R(K)>1$ means that spend has grown proportionally more than completion relative to one-shot serving. We call this increase the retry tax.

The verifier must determine success before the controller retries. If verification is delayed or unavailable, the stopping rule and its costs change. The complete agent configuration, including its prompt, tools, scaffold, and environment, remains fixed throughout the policy.

\subsection{Selection among unresolved tasks}
\label{sec:selection}

Let $P\in[0,1]$ denote the success probability of a randomly drawn task. A task's probability remains fixed across attempts. Conditional on $P$, outcomes are independent, and every attempt costs the same $c>0$. The tasks unresolved before attempt $r$ form its \emph{survivor pool}. Easier and harder refer to success probability for this fixed agent and verifier, rather than an intrinsic task label.

In the introduction's equal mixture of tasks with success probabilities 0.9 and 0.1, one attempt completes half the incoming tasks. Only 10 percent of the failures come from the easier group. A second attempt raises completion to 0.59 at a mean spend of $1.5c$ per incoming task, so cost per completion rises from $2c$ to $1.5c/0.59$. If every task instead has success probability 0.5, the same one-shot mean, two attempts complete 0.75 of tasks for the same $1.5c$ spend, leaving cost per completion at $2c$. This same-mean homogeneous control isolates the role of task heterogeneity: additional completion alone need not raise cost per completion.

Write
\begin{equation}
G_K(P)=\sum_{r=0}^{K-1}(1-P)^r.
\end{equation}
For a task with success probability $P$, $(1-P)^r$ is the probability of reaching attempt $r+1$, because the first $r$ attempts must all fail. Summing these probabilities gives the expected number of attempts reached before success or the cap, $G_K(P)$. Since $P G_K(P)=1-(1-P)^K$,
\begin{align}
Q(K)&=\E[P G_K(P)], & B(K)&=c\E[G_K(P)], \\
E(1)&=\frac{c}{\E[P]}, &
R(K)&=\frac{\E[P]\E[G_K(P)]}{\E[P G_K(P)]}.
\end{align}
Tasks with smaller $P$ remain eligible for more attempts. Rearranging the covariance definition makes this association explicit:
\begin{equation}
\label{eq:covariance}
R(K)=1-\frac{\Cov(P,G_K(P))}{Q(K)}.
\end{equation}

\begin{proposition}[Retry tax under task heterogeneity]
\label{prop:retry-tax}
Under the stationary, conditionally independent, constant-cost model, if $\E[P]>0$, then $R(K)\geq 1$ for every supported cap $K$. For $K\geq 2$, equality holds if and only if $P$ is constant almost surely.
\end{proposition}

\begin{proof}
For $K\geq 2$, $G_K(P)$ is strictly decreasing in $P$. Thus $\Cov(P,G_K(P))\leq 0$, with equality only when $P$ is constant almost surely. Equation~\ref{eq:covariance} gives the result. The $K=1$ case holds by definition.
\end{proof}

The same-mean homogeneous control explains the equality case. If every task has probability $p>0$, completion rises from $p$ to $1-(1-p)^K$, and stopped spend rises in the same proportion. Hence $E(K)=c/p=E(1)$ at every cap. Under the proposition's assumptions, it is heterogeneity between tasks that makes $R(K)>1$ for $K\geq2$.

At cap two, $G_2(P)=2-P$ gives an exact expression for the excess cost ratio.
\begin{corollary}[Cap-two variance identity]
\label{cor:cap-two}
\begin{equation}
\label{eq:variance}
R(2)=1+\frac{\Var(P)}{Q(2)}.
\end{equation}
\end{corollary}
Holding one-shot mean success fixed, the cap-two excess is determined by between-task variance. Feedback, drift across attempts, and task-dependent costs can change this relationship; they are outside this constant-cost identity.

\subsection{Marginal completion cost and cap choice}

An operator choosing a cap needs the completions and spend added by another attempt, rather than only the average over all completions. For a positive additional completion probability, define the marginal completion cost as
\begin{equation}
\Delta B_r=B(r)-B(r-1),\quad
\Delta Q_r=Q(r)-Q(r-1),\quad
M_r=\frac{\Delta B_r}{\Delta Q_r},
\end{equation}
with $B(0)=Q(0)=0$. Under the constant-cost model,
\begin{equation}
\label{eq:marginal}
M_r=
\frac{c\E[(1-P)^{r-1}]}
     {\E[P(1-P)^{r-1}]}.
\end{equation}
The numerator is the additional spend per incoming task when attempt $r$ is allowed, and the denominator is the completion probability it adds. Equivalently, $M_r=c/\mu_r$, where
\begin{equation}
\mu_r=\frac{\E[P(1-P)^{r-1}]}{\E[(1-P)^{r-1}]}
\end{equation}
is mean success probability among tasks reaching that attempt.

\begin{samepage}
If $\Var_r(P)$ is variance in this survivor pool, a further failure changes its mean according to
\begin{equation}
\label{eq:survivor-recurrence}
\mu_{r+1}=\mu_r-\frac{\Var_r(P)}{1-\mu_r}.
\end{equation}
This recurrence expresses the selection in the 90/10 example at every attempt.

\begin{proposition}[Monotone marginal cost]
\label{prop:marginal-cost}
Under the assumptions of Proposition~\ref{prop:retry-tax}, $M_{r+1}\geq M_r$ whenever both marginal completion increments are positive. Equality holds if and only if the survivor pool reaching attempt $r$ is homogeneous in $P$.
\end{proposition}

\begin{proof}
Equation~\ref{eq:survivor-recurrence} shows that $\mu_{r+1}\leq\mu_r$, with equality exactly when $\Var_r(P)=0$. Since $M_r=c/\mu_r$, the claim follows.
\end{proof}
\end{samepage}

Suppose each completion has the same value $V$, in the same units as spend. Expected additive utility is $U(K)=VQ(K)-B(K)$. When $\Delta Q_r>0$, allowing attempt $r$ adds enough completion value to cover its spend exactly when $V\geq M_r$. If completion increments are positive and marginal costs are nondecreasing, as in Proposition~\ref{prop:marginal-cost}, once an increment costs more than $V$ per added completion, every later increment does too. Adjacent comparisons therefore identify the best common cap, with equality allowing either neighboring cap. A zero completion increment instead has $\Delta U_r=-\Delta B_r$ and must be evaluated directly.

If marginal costs are nonmonotone, an unfavorable increment can precede a preferable later cap, so the operator must compare $VQ(K)-B(K)$ over all supported caps. Completion value depends on the application and may differ across tasks. The common-cap comparison above also differs from choosing which individual task to retry, which requires task-specific information and an allocation policy.

Table~\ref{tab:metric-questions} summarizes the questions answered by the average and marginal quantities.

\begin{table}[t]
  \caption{Retry quantities and the questions they answer.}
  \label{tab:metric-questions}
  \small
  \begin{tabular}{@{}lp{0.70\columnwidth}@{}}
    \toprule
    Quantity & Question answered \\
    \midrule
    $Q(K)$ & What share of incoming tasks completes within the cap? \\
    $B(K)$ & How much does the stopped service spend per incoming task? \\
    $E(K)$ & How much does it spend per completion? \\
    $R(K)$ & How does cost per completion compare with one-shot serving? \\
    $M_r$ & What does attempt $r$ cost per completion that it adds? \\
    \bottomrule
  \end{tabular}
\end{table}

\subsection{When task costs also differ}

A task can be both hard to complete and expensive to attempt. Let $Z$ be a latent task state, with success probability $P=P(Z)$ and mean attempt cost $C=C(Z)>0$. Suppose $S_r\mid Z$ are independent and identically distributed Bernoulli$(P(Z))$ outcomes and, for reached attempts,
\begin{equation}
\E[C_r\mid Z,S_{1:r-1},C_{1:r-1}]=C(Z),
\end{equation}
with finite $\E[C]$ and positive completion probability. Current cost may depend on current success, but its conditional mean before observing that outcome must remain unchanged after earlier failures. Then
\begin{equation}
\label{eq:variablecost}
R(K)=
\frac{1+\Cov(C,G_K)/(\E[C]\E[G_K])}
     {1+\Cov(P,G_K)/(\E[P]\E[G_K])}.
\end{equation}
The denominator describes the selection due to success heterogeneity. The numerator describes whether tasks that remain for more attempts also have higher mean attempt costs. Costly hard tasks can increase the ratio; a different cost-success association can reduce it.

Equation~\ref{eq:variablecost} separates the two associations under the stated conditional-mean model. Repeated-run records offer a complementary way to calculate completion and spending: average the stopped sequences formed from their observed outcome-cost pairs. This finite-record calculation describes the available attempts directly, while the decomposition describes a process satisfying the model assumptions.

\section{Measuring the Trade-off from Repeated Runs}
\label{sec:method}

\subsection{Completion and spend over recorded orders}

Repeated benchmark runs give several observed outcome-cost pairs for each task. For cap $K$, each eligible order consists of $K$ distinct recorded attempts. We stop each order at success or the cap and average over all eligible orders, first within a task and then over equally weighted tasks. This \emph{ordered-subset replay} gives the exact finite-record average of completion and stopped recorded spend. For example, with attempts A, B, and C and cap two, it averages every ordered pair and charges the second member only if the first failed. Each cost stays with its recorded outcome.

The average describes randomized sampling without replacement from the observed panel. Interpreting it as completion and spend for fresh retries requires the records and the target retry process to share the same joint outcome-cost law over the relevant attempts. Appendix~\ref{app:fresh-process} gives sufficient conditions and the distinction from feedback-conditioned evaluation.

At constant attempt cost, completion at every earlier cap already determines stopped spend: an attempt is paid for exactly when the preceding attempts have not completed the task. For task $i$, with replay completion $\bar q_i(K)$ and spend $\bar b_i(K)$,
\begin{equation}
\bar b_i(K)=c\sum_{r=0}^{K-1}[1-\bar q_i(r)],\qquad \bar q_i(0)=0.
\label{eq:prefix-identity}
\end{equation}
The completion term is the standard finite-sample combinatorial pass@k estimator~\cite{chen2021evaluating}. With variable costs, the stopped calculation also needs the costs paid at the eligible positions. Appendix~\ref{app:replay} gives the exact replay formulas.

\subsection{Three fixed panels}
\label{sec:design}

The primary panel comes from released artifacts of \emph{On Randomness in Agentic Evals}~\cite{bjarnason2026randomness}: 12 fixed configurations, each with ten attempts on the same 500 SWE-bench Verified tasks. Configurations cross Nano-agent and R2E-Gym, the DeepSWE-Preview, Devstral, and Qwen3 coding models, and released default and temperature-zero decoding settings. We assign unit cost to each attempt. Every released status other than resolved is a failure. The analysis uses task identifiers and binary statuses from files under CC BY 4.0, excluding prompts, trajectories, patches, and messages.

The $\tau^2$-bench panel has four trials per task for four fixed configurations on 164 airline and telecom tasks. The benchmark evaluates dual-control conversational agents~\cite{barres2026tau2}, extending the tool-agent-user setting of $\tau$-bench~\cite{yao2025taubench}. The OpenHands panel has four runs on 500 SWE-bench Verified tasks for one fixed CodeActAgent v0.62.0 configuration using gpt-5-2025-08-07, with attempt costs released for a study of coding-agent token consumption~\cite{bai2026agentspend}. Both monetary panels retain costs as recorded, rather than reconstructing bills. We treat this reanalysis of previously inspected benchmark records as exploratory.

\subsection{Endpoint summaries and uncertainty}

Figure~\ref{fig:retry-tradeoff} and Appendix~\ref{app:cap-trajectories} show completion and spend at every supported cap. The plots use point estimates, including marginal prices, without interval bars. All three panels record at least four attempts per task, so we compare four attempts with one: within each configuration, $R_4=(B(4)/Q(4))/(B(1)/Q(1))$. To summarize a panel, we take an equally weighted median of the configurations' $\log R_4$ values and exponentiate. With an even number of configurations, the median log is the arithmetic mean of the two central ordered logs, so the displayed ratio is their geometric midpoint.

For the four-versus-one comparisons and the cost replacements in Appendix~\ref{sec:cost-comparisons}, 20,000 shared-task bootstrap draws give nominal pointwise 95 percent percentile intervals, recomputing the complete statistic in each draw. Each draw uses the same sampled tasks across attempts, configurations, and cost calculations, preserving their dependence. In $\tau^2$-bench, sampling stays within the airline and telecom domains to retain their fixed composition. These intervals describe task resampling within the evaluated frames and fixed configurations. We separately examine whether every bootstrap draw gives a finite final marginal-cost estimate. For the recorded-run endpoint and contrast intervals and these separately examined marginal intervals, we leave an interval unavailable if any planned draw is nonfinite, retaining every draw rather than computing the interval from finite draws alone.

\section{Results}
\label{sec:results}

\subsection{What each cap buys}

\begin{figure*}[t]
  \centering
  \includegraphics[width=\textwidth]{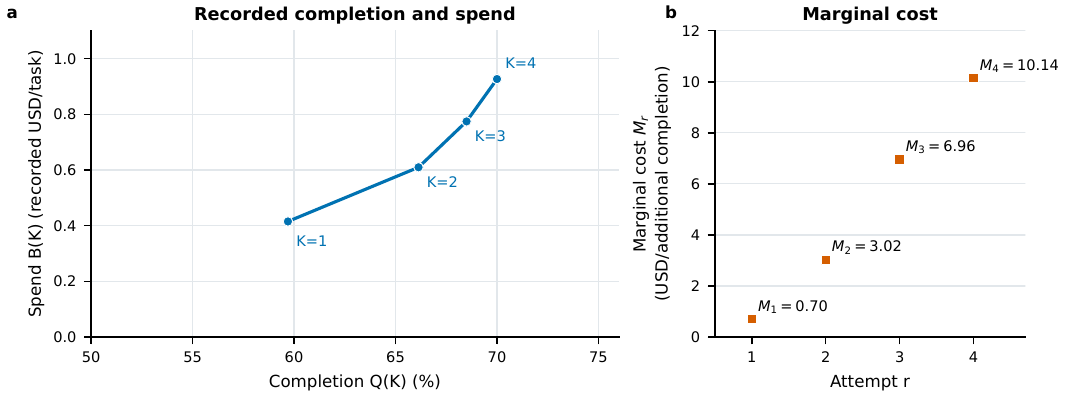}
  \caption{OpenHands, $n=500$ incoming tasks. Left: completion $Q(K)$ (\%) and stopped spend $B(K)$ (recorded USD/task); segments join adjacent discrete caps. Right: marginal cost $M_r$ (USD/additional completion). Point estimates; no uncertainty intervals. The completion axis is focused. $M_1=E(1)$ is the one-shot cost per completion. Marginal costs are calculated from unrounded estimates.}
  \Description{The first panel marks completion and stopped spending at each of the four OpenHands retry caps. The second marks the corresponding marginal cost per additional completion at each attempt.}
  \label{fig:retry-tradeoff}
\end{figure*}

Figure~\ref{fig:retry-tradeoff} follows one OpenHands configuration through all four supported caps. Completion rises from 59.7 percent to 70.0 percent while mean stopped recorded spend rises from \$0.4152 to \$0.9266 per incoming task. At the intermediate caps, completion is 66.1 percent at cap two and 68.5 percent at cap three, at mean spend of \$0.6098 and \$0.7745, respectively. Each pair of neighboring points in panel (a) gives the added spend and completion gain. Dividing the added spend by the gain in completion probability, expressed as a fraction, gives the adjacent marginal cost $M_r$ in panel (b).

The average cost per completion rises from \$0.6954 at cap one to \$1.3237 at cap four, but the latter average conceals the price of the last completions. The adjacent marginal costs are about \$3.02, \$6.96, and \$10.14 per additional completion for attempts two, three, and four. Combining those later increments gives $(B(4)-B(1))/(Q(4)-Q(1))$, about \$4.97 per additional completion from cap one to four. That combined price averages over all completions added after the first attempt; $M_4$ prices only the last increment.

Appendix~\ref{app:cap-trajectories} gives the corresponding trajectories for every primary and $\tau^2$-bench configuration, through ten and four caps, respectively. At any pair of caps, the additional completion and spending can be compared with the application's completion value.

\subsection{Four attempts versus one across panels}

\begin{table}[t]
\centering
\caption{Cost per completion at four attempts relative to one.}
\label{tab:h4-panels}
\begin{tabular}{lrrrr}
\toprule
Panel & Tasks & Configs. & \shortstack{Cost ratio\\($R_4$)} & \shortstack{95\%\\interval} \\
\midrule
Primary & 500 & 12 & 1.7731 & [1.7096, 1.8476] \\
$\tau^2$ & 164 & 4 & 1.2365 & [1.1500, 1.3309] \\
OpenHands & 500 & 1 & 1.9036 & [1.7860, 2.0232] \\
\bottomrule
\end{tabular}
\par\smallskip
\begin{minipage}{\linewidth}\footnotesize Estimate is $\exp(\operatorname{median}_c\log R_{4,c})$, where $R_4=(B_4/Q_4)/(B_1/Q_1)$, $B$ is mean stopped spend, and $Q$ is completion probability. Brackets are nominal pointwise 95\% percentile intervals from 20,000 shared-task bootstrap draws. Tasks are resampling units; configurations and $\tau^2$ domain composition are fixed.\end{minipage}
\end{table}

At cap four, the summaries in Table~\ref{tab:h4-panels} correspond to about 77 percent higher cost per completion in the primary panel, 24 percent in $\tau^2$-bench, and 90 percent in OpenHands, relative to one attempt. The first two summarize ratios within fixed configurations; they do not divide costs pooled across configurations or imply that every configuration has the same increase. The OpenHands summary is the single configuration in Figure~\ref{fig:retry-tradeoff}.

In the unit-cost primary panel, replacing attempt costs by task or configuration means leaves replay unchanged. Replay also reproduces Equation~\ref{eq:prefix-identity} up to floating-point error. These equalities check the accounting implementation; Appendix~\ref{sec:cost-comparisons} describes the replacements.

\subsection{Supporting checks on cost and uncertainty}

\paragraph{Retaining costs with outcomes.}
We keep outcomes fixed and compare recorded costs with costs set to each task's mean. The replacement assigns the same cost to successes and failures on a task while preserving differences in mean cost between tasks. Relative to task-mean costs, retaining the recorded costs gives a positive summary change in $R_4$: about 0.54 percent for $\tau^2$-bench, with a nominal pointwise 95 percent interval of [0.24, 0.86] percent, and 0.38 percent [0.14, 0.64] for OpenHands. These are relative changes in the retry-to-one-shot cost ratio, not percentage-point changes in completion. The $\tau^2$ summary exponentiates the median of the paired, within-configuration log changes; it differs from subtracting the two panel median logs. Both summary changes are below one percent, and the configuration-specific directions are mixed, including a negative estimate and intervals spanning zero. Retaining within-task pairing therefore changes the average-cost comparison only slightly in these panels. Appendix~\ref{sec:cost-comparisons} separates these two summaries and reports the other cost replacements.

\paragraph{Support for marginal intervals.}
The plots display marginal point estimates; a separate check examines whether resampled tasks yield finite estimates at the final supported cap, where few tasks may contribute additional completions. For example, for a task with exactly four recorded attempts, consider raising the cap from three to four. This adds a completion only if the first three attempts in the replayed order fail and the fourth succeeds. Because the order uses all four distinct records without replacement, only tasks with exactly one recorded success can satisfy this condition, regardless of their costs. A task-bootstrap draw containing no such tasks therefore has a zero final completion increment and an undefined marginal cost.

In the monetary $\tau^2$-bench panel, some bootstrap draws have nonfinite $M_4$, so the affected intervals are unavailable under the reporting rule in Section~\ref{sec:method} even though all point estimates are available. The nonfinite-draw counts alone do not establish the cause of each failure. Appendix~\ref{app:increment-support} gives those counts and derives the final increment for a general recorded repeat count.

\paragraph{Checking interval coverage.}
Do nominal 95 percent intervals contain known truth at the advertised rate? A separate synthetic-data experiment generates 64 independent datasets under each of five fixed designs, then bootstraps tasks within each dataset. In three primary-style designs, including a high-success homogeneous control, percentile and bias-corrected and accelerated (BCa) intervals target the median of configuration $\log R_4$ values; in one OpenHands-style design, they target the single configuration's $\log R_4$. In the $\tau^2$-style design, standardized-maximum intervals target the four configuration $\log R_4$ values, with joint containment requiring all four intervals to contain their targets.

The BCa interval for the high-success homogeneous design's median log ratio contains its target in 54/64 datasets. The $\tau^2$-style intervals jointly contain all four targets in 54/64 datasets as well. Appendix~\ref{app:synthetic} reports all 13 combinations of design, interval method, and target, with the interval constructions, delivery results, and Monte Carlo bounds. These fixed-design results check specified targets; they do not estimate coverage of the historical intervals.

\section{Related Work}

\subsection{Repeated attempts and their costs}

Repeated attempts can be assessed in different ways. pass@k measures at least one success among $k$ samples and has a combinatorial estimator~\cite{chen2021evaluating}; pass$^k$ measures success on every attempt~\cite{bjarnason2026randomness}. Agent evaluations include independent repeated executions~\cite{bjarnason2026randomness} and retry protocols that supply targeted feedback after failure~\cite{gonzalez2026reliability}. Inference-time scaling studies relate repeated-sampling completion to compute and the distribution of task success probabilities~\cite{schaeffer2025monkeys}. For stop-on-success serving, spending further depends on which attempts are reached.

Cost-of-Pass defines expected inference cost per successful solution under independent repeated attempts~\cite{erol2026costofpass}. HYPE-EDIT-1 expresses capped expected attempts and effective cost per successful edit through task pass rates~\cite{chan2026hypeedit}. We use these cost quantities to analyze the changing unresolved task mix and to compare average cost with the marginal price of completion. At constant cost, the complete prefix completion curve determines spend exactly; variable-cost records also require their outcome-cost pairs.

Murphy's working paper evaluates economic restart policies by ordered replay, includes verifier and fallback prices, and uses cross-fitting for selected-policy evaluation~\cite{murphy2026restart}. Seq@K compares independent and sequential calibration for held-out model selection and reports stopped-cost frontiers under feedback~\cite{yen2026seqk}. Here the comparison is across retry caps for each fixed configuration. Relating its completion trajectory to stopped spending distinguishes the average cost of all completions from the marginal cost needed to assess a larger cap.

Repeated coding-agent runs can differ substantially in token consumption within the same task~\cite{bai2026agentspend}, and cost-aware benchmarks compare complete configurations~\cite{chi2026agentmeter}. Our retry-to-one-shot ratios instead compare caps within a configuration. Cost replacements distinguish within-task pairing from the additional removal of between-task cost structure. Contextual-bandit replay retains randomized events whose logged actions match a target action~\cite{li2011offline}; here the records contain repeated outcome-cost pairs on the same task and no action propensities. Their finite-order average has a deployment interpretation only under the joint-process condition in Section~\ref{sec:method}.

\subsection{Heterogeneity and survivor selection}

Frailty and mixed-hazard models describe how populations change as members with different exit probabilities leave at different rates~\cite{vaupel1985ruses,proschan1963dfr,alvarez2024duration}. Discrete-time duration models also estimate duration dependence while allowing unrestricted unobserved heterogeneity~\cite{alvarez2025prices}. Section~\ref{sec:theory} specializes this mechanism to static retries, relating the survivor mix to cost per completion through the covariance and cap-two identities. Its assumptions distinguish that account from feedback-conditioned or otherwise adaptive retries.

\section{Discussion and Limitations}

\paragraph{Using a cap comparison.}
Reporting completion $Q(K)$, stopped spend $B(K)$, and cost per completion $E(K)$ together lets an operator compare the completion achieved at each cap with the spending it requires. $R(K)$ states the average-cost change relative to one-shot serving. For a larger cap, $\Delta Q_r$, $\Delta B_r$, and $M_r$ instead answer how much extra completion it buys and at what additional cost. The OpenHands trajectory shows why an average such as $E(4)$ cannot substitute for the final marginal cost. A high marginal cost can still be acceptable when those completions are valuable.

\paragraph{What the records establish.}
The empirical estimates describe eligible orders of the available runs, with task resampling for the evaluated configurations and task frames. They do not estimate variation across a population of agents or unseen repositories. The primary and OpenHands panels also use the same benchmark task frame, so the three panels are not three independent task populations. Reanalyzing these outcomes supplies no independent validation on new data. Retaining each task's attempts and configurations together in resampling preserves their dependence; treating attempts as independent uncertainty units could overstate effective sample size.

\paragraph{Connecting replay to a retry service.}
The finite-record calculation identifies completion and stopped spend through the observed repeat count. The model propositions describe a stationary, conditionally independent process with a common attempt cost, and need not govern the monetary replay curves. A fresh-retry interpretation requires the same joint outcome-cost process as the records. Informative missingness, outcome-based run selection, provider drift, and adaptive responses can break that connection. Feedback-conditioned policies require chronological outcome-cost records or a transition model to evaluate their sequential decisions.

\paragraph{Verification and prices.}
The stopping rule assumes that the verifier supplies an outcome after each attempt. Delayed, noisy, or partial verification changes which attempts are reached, and benchmark acceptance need not equal operational correctness. Monetary estimates inherit the released accounting conventions, possible record errors, and prices at recording time. Changing recorded costs measures an accounting contrast; a causal economic loss or regret comparison would require an identified intervention and decision policy.

\paragraph{Uncertainty.}
The synthetic experiments identify two interval-coverage shortfalls under their specified laws. Their Monte Carlo bounds concern containment and delivery under those laws. With the fixed designs and numbers of simulated datasets used here, broader calibration claims or a ranking of interval methods would need separate validation.

\section{Conclusion}

Consider static retries with stationary task success probabilities, conditionally independent attempts, and a common attempt cost. When success probabilities differ across tasks, easier tasks tend to succeed earlier, leaving a harder unresolved pool. The resulting selection makes average cost per completion at least as high as in one-shot serving and makes marginal cost per additional completion nondecreasing wherever the completion increments are positive. A higher terminal completion rate therefore leaves a spending question: what do the later completions cost?

The replay estimates show how average and marginal costs differ. For the OpenHands configuration, estimated completion is 59.7 percent with one attempt and 70.0 percent with four, while average cost per completion rises from about \$0.70 to \$1.32. Raising the cap from three to four costs about \$10.14 per additional completion. That last price answers a different question from the average across all completed tasks. A cap comparison becomes an economic decision when it asks whether the additional completions justify the additional spending.

\section*{Ethical Considerations}

This study reanalyzes public benchmark records and simulated data, with no new experimental model or judge calls, participants, or real-world actions. AI tools assisted analysis-code development, manuscript drafting, and consistency checks. The licensed primary-panel analysis uses task identifiers and binary statuses rather than prompts, messages, trajectories, or patches. Analysis provenance retains source records, licenses, versions, and hashes. The recorded source inspection did not establish an explicit redistribution license for the OpenHands trajectory archive released by Bai et al.~\cite{bai2026agentspend}, so permission to redistribute derived panels remains unresolved.

\begin{figure*}[t]
  \centering
  \includegraphics[width=\textwidth]{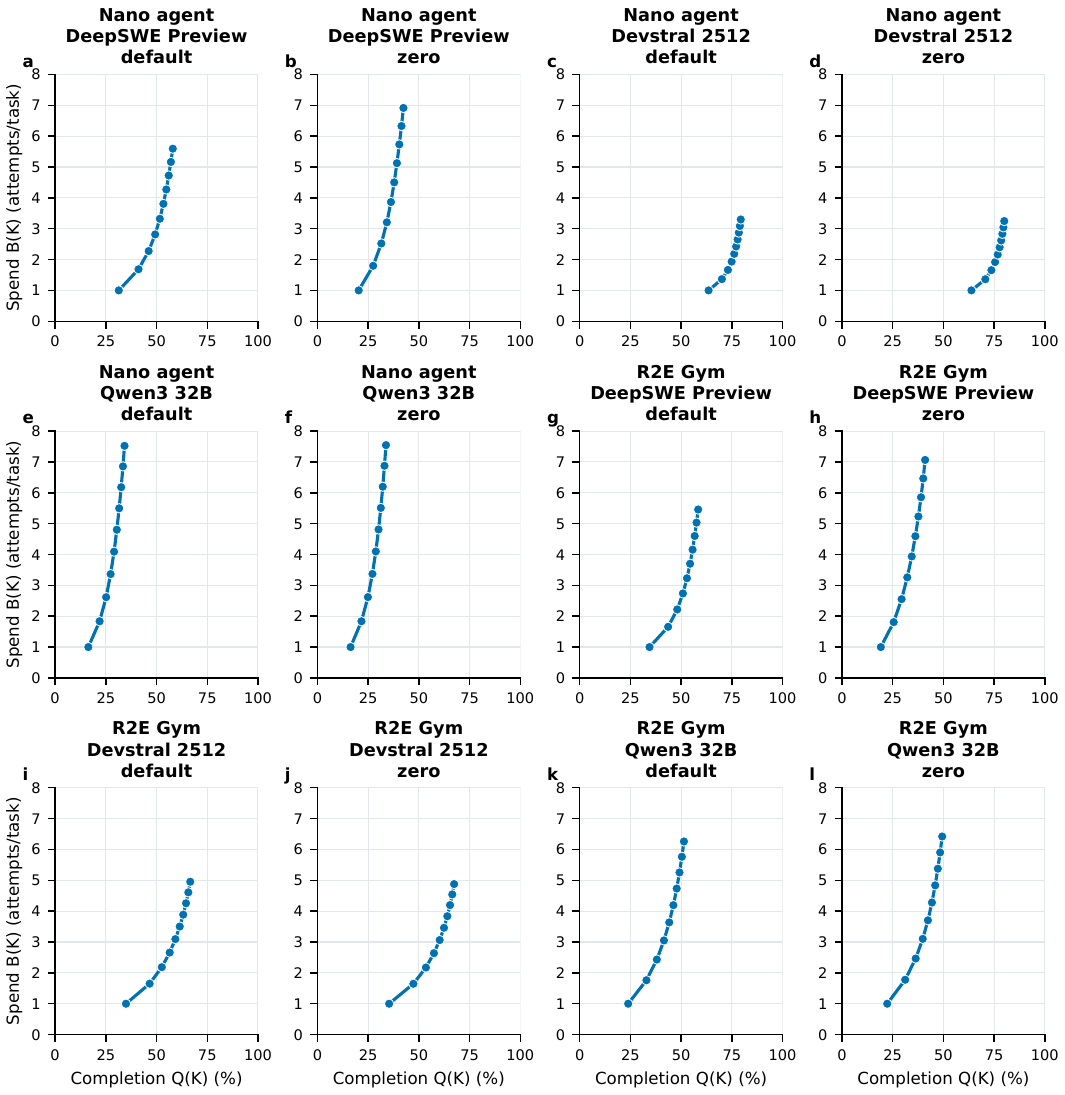}
  \caption{All 12 primary configurations, $n=500$ incoming tasks each. Completion $Q(K)$ (\%) versus stopped spend $B(K)$ (attempts/task). Segments join adjacent discrete caps $K=1,\ldots,10$. Point estimates; no uncertainty intervals.}
  \Description{Twelve configuration-specific panels show completion and mean stopped unit-cost attempts at every supported cap in the primary panel.}
  \label{fig:primary-caps}
\end{figure*}

\appendix

\section{All Supported Cap Trajectories}
\label{app:cap-trajectories}

\begin{figure*}[t]
  \centering
  \includegraphics[width=\textwidth]{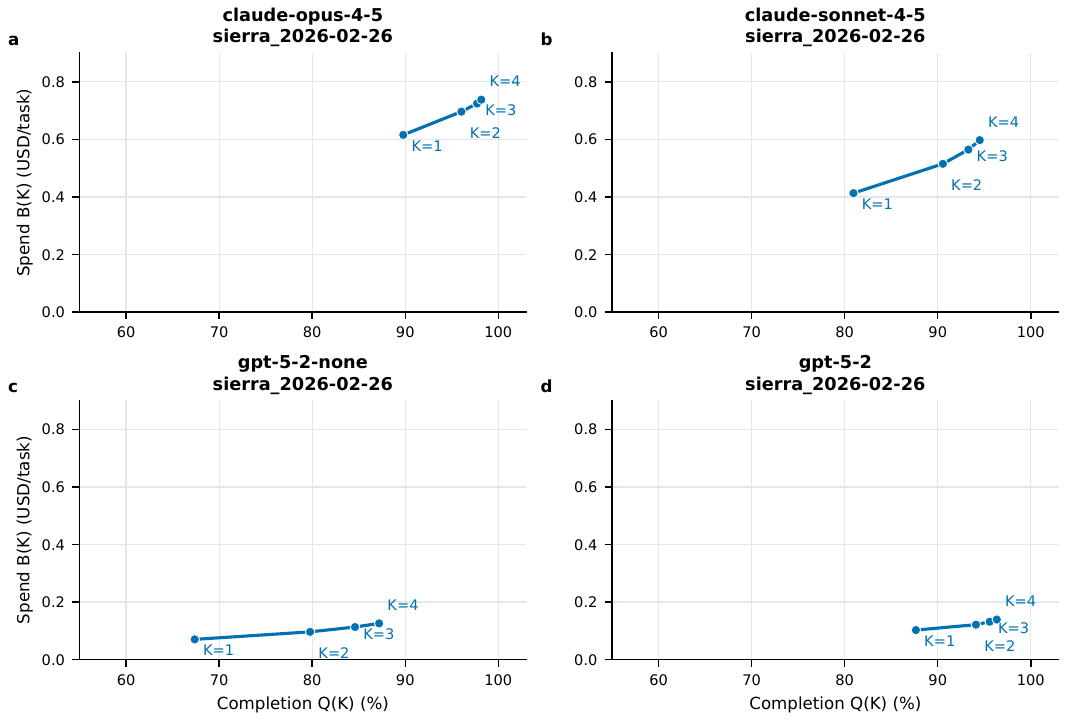}
  \caption{All four $\tau^2$ configurations, $n=164$ incoming tasks each. Completion $Q(K)$ (\%) versus stopped spend $B(K)$ (recorded USD/task). Segments join adjacent discrete caps $K=1,\ldots,4$. Point estimates; no uncertainty intervals. Completion axes are focused.}
  \Description{Four configuration-specific panels show completion and mean stopped monetary spend at every supported cap in the tau-squared benchmark panel.}
  \label{fig:tau-caps}
\end{figure*}

Figures~\ref{fig:primary-caps} and~\ref{fig:tau-caps} show every fixed configuration and supported cap in the primary and $\tau^2$-bench panels. Together with Figure~\ref{fig:retry-tradeoff}, they describe the completion-spend trade-off in all three panels. Each point is a discrete cap, and straight connections mark adjacent caps only. The primary panel measures spend in unit-cost attempts; the monetary panels use recorded USD. The trajectories show point estimates with no interval bars.

\section{Exact Replay and Finite-Sample Targets}
\label{app:replay}

\subsection{Averaging stopped orders}

Suppose task $i$ has $m\ge K$ recorded attempts with outcomes $S_{ij}$ and costs $C_{ij}$. Let $\mathcal{P}(m,K)$ contain all ordered length-$K$ tuples of distinct attempt indices, so its size is $(m)_K=m!/(m-K)!$. For tuple $a=(a_1,\ldots,a_K)$, completion and stopped spend are
\begin{align}
q_i(a)&=1-\prod_{r=1}^{K}(1-S_{i a_r}), \\
b_i(a)&=\sum_{r=1}^{K}C_{i a_r}\prod_{s<r}(1-S_{i a_s}).
\end{align}
Averaging those orders gives the task's replay values,
\begin{equation}
\bar q_i(K)=\frac{1}{(m)_K}\sum_{a\in\mathcal{P}(m,K)}q_i(a),\quad
\bar b_i(K)=\frac{1}{(m)_K}\sum_{a\in\mathcal{P}(m,K)}b_i(a).
\end{equation}
Pooling $n$ equally weighted incoming tasks gives
\begin{equation}
\widehat Q(K)=\frac{1}{n}\sum_i\bar q_i(K),\quad
\widehat B(K)=\frac{1}{n}\sum_i\bar b_i(K),\quad
\widehat E(K)=\frac{\widehat B(K)}{\widehat Q(K)}.
\end{equation}
Each task contributes its full outcome-cost record before task averaging. In particular, a costly attempt is paid only when earlier attempts in the selected order failed; separately averaging costs and success rates removes that relationship.

The averages do not require enumerating all $(m)_K$ tuples. If $F_i=\sum_j(1-S_{ij})$ counts task $i$'s failures and $(x)_t$ denotes a falling factorial, then
\begin{align}
\bar q_i(K)&=1-\frac{(F_i)_K}{(m)_K}, \\
\bar b_i(K)&=\sum_{r=1}^{K}\frac{1}{m}\sum_{j=1}^{m} C_{ij}
\frac{(F_i-1+S_{ij})_{r-1}}{(m-1)_{r-1}}.
\label{eq:closed-form-replay}
\end{align}
To read the spend expression, reserve attempt $j$ for position $r$. The remaining $m-1$ records contain $F_i-1$ failures if $j$ failed, or $F_i$ failures if it succeeded, giving $F_i-1+S_{ij}$ in either case. The inner factor is therefore the probability that all $r-1$ earlier attempts failed, conditional on attempt $j$ occupying position $r$. The calculation takes $O(mK)$ operations and identifies caps only through the recorded repeat count $m$. Writing $X_i=\sum_jS_{ij}$, its completion term $1-(m-X_i)_K/(m)_K$ is the combinatorial pass@k estimator~\cite{chen2021evaluating}.

\subsection{Support for the final increment}
\label{app:increment-support}

At the final supported cap $K=m$, a replayed order uses all $m$ distinct records. A task gains a completion over cap $m-1$ only when its sole recorded success occupies the last position, which occurs in a fraction $1/m$ of orders. Tasks with no successes cannot gain a completion, and tasks with two or more successes must already have completed before the last position. This condition concerns outcomes alone and does not depend on costs.

Let $n_0$ and $n_1$ count tasks with zero and exactly one success among their $m$ recorded attempts. Every zero-success task reaches the last attempt, whereas a one-success task reaches it only when its success is last; tasks with two or more successes never reach it. With each attempt assigned unit cost, the increments at $K=m$ are $\Delta Q_m=n_1/(mn)$ and $\Delta B_m=n_0/n+n_1/(mn)$, so $M_m=1+m n_0/n_1$ when $n_1>0$. If $n_1=0$, the last attempt adds no completions and its marginal cost is undefined, including in any task-bootstrap draw with no one-success tasks.

In $\tau^2$-bench at cap four, realized-cost $M_4$ is nonfinite in 905, 5, 0, and 113 of the 20,000 bootstrap draws for c01 through c04 in Table~\ref{tab:h4-pairing}'s fixed configuration order. All $M_4$ point estimates remain available. Under the reporting rule in Section~\ref{sec:method}, the affected marginal intervals are unavailable and all draws are retained. The counts describe the draws that fail to give a finite marginal estimate.

\subsection{Interpreting fresh attempts}
\label{app:fresh-process}

Conditionally independent and identically distributed attempts from the target process are sufficient for the replay components to estimate its completion and stopped spend. An exchangeable process is also sufficient if it extends to future attempts and has the same law over the relevant $K$ attempts. Exchangeability within the observed panel alone does not establish that connection. Resetting the prompt and workspace specifies the request but does not establish a match to the target outcome-cost process or show that provider caches have been cleared. Feedback-conditioned retries instead require chronological evaluation or a sequential model.

\subsection{Independent-attempt plug-in calculation}

Under conditionally independent and identically distributed sampling, the component estimates of $Q(K)$ and $B(K)$ are symmetric U-statistics and unbiased for the corresponding policy quantities~\cite{hoeffding1948class}. Substituting the empirical task success rate $\widehat p_i=X_i/m$ into the stationary formulas answers a different finite-sample question: for $K\geq2$, concavity makes $1-(1-\widehat p_i)^K$ downward biased, while under unit cost and $K\geq3$, convexity makes $G_K(\widehat p_i)$ upward biased. Even with unbiased component estimates, the quotient $\widehat E(K)$ is a ratio estimator and is not exactly unbiased.

\section{Cost Replacements and All Contrasts}
\label{sec:cost-comparisons}
\label{sec:appendix-evidence}

The fixed $\tau^2$-bench configurations are:
\begin{description}
  \item[c01] claude-opus-4-5\_sierra\_2026-02-26
  \item[c02] claude-sonnet-4-5\_sierra\_2026-02-26
  \item[c03] gpt-5-2-none\_sierra\_2026-02-26
  \item[c04] gpt-5-2\_sierra\_2026-02-26
\end{description}
This order is used in Table~\ref{tab:h4-pairing} and the nonfinite-draw counts in Appendix~\ref{app:increment-support}.

The comparisons hold tasks and outcomes fixed while changing how attempt cost enters replay. \emph{Realized-cost replay} retains each attempt's recorded cost. \emph{Task-mean replay} replaces every attempt's cost by that task's arithmetic mean, preserving between-task cost differences while removing within-task cost-outcome alignment. A positive realized-minus-task-mean contrast in log $R_4$ means retaining the pairing raises the recorded retry-to-one-shot ratio. 

\emph{Global-mean replay} replaces every cost by the configuration's mean, removing between-task cost structure as well. The \emph{task-success plug-in} procedure instead uses $\widehat p_i=X_i/m$, completion $1-(1-\widehat p_i)^K$, and task-mean cost times $G_K(\widehat p_i)$. This represents independent draws at empirical task success probabilities, whereas exact replay samples finite tuples without replacement. Contrasts involving the plug-in procedure therefore compare different targets rather than measuring known-truth bias.

The within-configuration change and the change in a panel summary are also distinct. For realized versus task-mean replay, define
\begin{equation}
\begin{aligned}
d_c &= \log R_{\mathrm{realized},c}(4)\\
    &\quad -\log R_{\mathrm{task\ mean},c}(4),\\
D &= \operatorname{median}_c d_c.
\end{aligned}
\end{equation}
Here $D$ is the median within-configuration change. Taking each procedure's panel median first and then subtracting gives the \emph{difference of panel median logs}. The two operations need not agree: separate medians can draw on different central configurations, so taking medians first need not retain the configuration pairing used in $d_c$. Both summaries use the same sampled tasks across procedures and configurations. With a single OpenHands configuration, the summaries coincide. All intervals below are nominal pointwise 95 percent percentile intervals from the 20,000 shared-task draws described in Section~\ref{sec:method}.

\begin{table*}[t]
\centering
\caption{Log-ratio contrasts under cost replacements and plug-in replay.}
\label{tab:h4-pairing}
\begin{minipage}[t]{0.32\textwidth}
\centering
\textit{Realized minus task mean}\par\smallskip
\begin{tabular}{@{}lr@{}}
\toprule
Target & \shortstack[r]{Contrast in $\log R_4$\\{}[95\% CI]} \\
\midrule
c01 & \shortstack[r]{0.0065\\{}[0.0027, 0.0107]} \\
c02 & \shortstack[r]{0.0095\\{}[0.0050, 0.0148]} \\
c03 & \shortstack[r]{0.0043\\{}[-0.0005, 0.0100]} \\
c04 & \shortstack[r]{-0.0004\\{}[-0.0061, 0.0055]} \\
$\tau^2\ D$ & \shortstack[r]{0.0054\\{}[0.0024, 0.0086]} \\
$\tau^2$ diff. & \shortstack[r]{0.0046\\{}[0.0006, 0.0088]} \\
OH $D$ & \shortstack[r]{0.0038\\{}[0.0014, 0.0064]} \\
\bottomrule
\end{tabular}
\end{minipage}
\hfill
\begin{minipage}[t]{0.32\textwidth}
\centering
\textit{Realized minus global mean}\par\smallskip
\begin{tabular}{@{}lr@{}}
\toprule
Target & \shortstack[r]{Contrast in $\log R_4$\\{}[95\% CI]} \\
\midrule
c01 & \shortstack[r]{0.0289\\{}[-0.0014, 0.0588]} \\
c02 & \shortstack[r]{0.0669\\{}[0.0318, 0.0999]} \\
c03 & \shortstack[r]{0.0636\\{}[0.0022, 0.1199]} \\
c04 & \shortstack[r]{0.1013\\{}[0.0573, 0.1459]} \\
$\tau^2\ D$ & \shortstack[r]{0.0652\\{}[0.0340, 0.1016]} \\
$\tau^2$ diff. & \shortstack[r]{0.0841\\{}[0.0530, 0.1167]} \\
OH $D$ & \shortstack[r]{0.0818\\{}[0.0524, 0.1102]} \\
\bottomrule
\end{tabular}
\end{minipage}
\hfill
\begin{minipage}[t]{0.32\textwidth}
\centering
\textit{Task mean minus plug-in}\par\smallskip
\begin{tabular}{@{}lr@{}}
\toprule
Target & \shortstack[r]{Contrast in $\log R_4$\\{}[95\% CI]} \\
\midrule
Primary $D$ & \shortstack[r]{-0.0536\\{}[-0.0574, -0.0484]} \\
Primary diff. & \shortstack[r]{-0.0538\\{}[-0.0602, -0.0483]} \\
c01 & \shortstack[r]{-0.0445\\{}[-0.0591, -0.0304]} \\
c02 & \shortstack[r]{-0.0669\\{}[-0.0859, -0.0489]} \\
c03 & \shortstack[r]{-0.0923\\{}[-0.1161, -0.0699]} \\
c04 & \shortstack[r]{-0.0436\\{}[-0.0606, -0.0282]} \\
$\tau^2\ D$ & \shortstack[r]{-0.0557\\{}[-0.0699, -0.0453]} \\
$\tau^2$ diff. & \shortstack[r]{-0.0552\\{}[-0.0691, -0.0434]} \\
OH $D$ & \shortstack[r]{-0.0513\\{}[-0.0642, -0.0393]} \\
\bottomrule
\end{tabular}
\end{minipage}
\par\smallskip
\begin{minipage}{\linewidth}\footnotesize All contrasts use log $R_4$ units. $D$ is the median of configuration-wise contrasts; diff. is the separate difference-of-medians estimand. They coincide for the single OpenHands configuration. $\tau^2$ uses 164 tasks; primary and OpenHands each use 500. Intervals are nominal pointwise 95\% percentile intervals from 20,000 shared-task bootstrap draws. c01 through c04 identify the fixed $\tau^2$ configurations listed in Appendix~\ref{sec:appendix-evidence}. Global mean also removes between-task cost structure; plug-in and exact replay have different targets.\end{minipage}
\end{table*}

In $\tau^2$-bench, realized-minus-task-mean $D$ is 0.005405 [0.002384, 0.008557], whereas the separate difference of panel median logs is 0.004573 [0.000590, 0.008785]. The single OpenHands contrast is 0.003792 [0.001408, 0.006357]. The configuration-specific $\tau^2$ rows include a negative estimate and intervals spanning zero, while both recorded-minus-task-mean panel summaries change $R_4$ by less than one percent. Global-mean contrasts remove more cost structure and are reported separately. The primary panel's unit costs make realized, task-mean, and global-mean replay agree, but its task-mean and plug-in targets still differ. Table~\ref{tab:h4-pairing} retains all configuration rows and all three comparison types.

\section{Synthetic Interval Containment}
\label{app:synthetic}

\subsection{Why use known targets?}

The historical targets are unknown, so those records alone cannot show how often their intervals contain truth. Five fixed synthetic designs instead generate datasets from specified laws, calculate their known targets, and compare intervals with those values. This exploratory check examines conditions relevant to the endpoint summaries, rather than fitting a model of a historical population.

\subsection{Designs and experimental units}

The designs vary shared task difficulty, ties among central true configuration ratios, and high success probabilities. P-\allowbreak SHARED-\allowbreak TIE shares easy and hard tasks across 12 configurations and has a tied pair of central true ratios. P-\allowbreak INDEPENDENT-\allowbreak SHAM preserves each configuration's marginal distribution while drawing difficulty independently across configurations. P-\allowbreak HIGH-\allowbreak HOMOGENEOUS is a high-success homogeneous control at constant cost. All three use 500 tasks and ten repeats per configuration. These choices define conditions for examining intervals; they are not established causes of any coverage shortfall.

The monetary designs vary task costs and allow current cost to depend on current outcome. M-\allowbreak TAU-\allowbreak HIGH has four configurations, 164 tasks in fixed airline and telecom domains, and four repeats. M-\allowbreak OPENHANDS-\allowbreak HIGH has one configuration, 500 tasks, and four repeats. Every design uses conditionally independent and identically distributed attempt vectors given task state, with no feedback or serial drift. Each design generates 64 independent simulated datasets. A dataset is the experimental unit, and methods are paired within it. Section~\ref{app:synthetic-laws} gives all probability and cost laws.

\subsection{Interval constructions}

The three P designs and M-\allowbreak OPENHANDS-\allowbreak HIGH use percentile and bias-corrected and accelerated (BCa) intervals. Each shared-task draw recomputes the complete median log-ratio statistic, or the scalar log ratio for the single OpenHands-style configuration. These four designs use 1,999 bootstrap draws within each simulated dataset. Each BCa jackknife value deletes one whole task across all configurations.

The $\tau^2$-style check uses 20,000 bootstrap draws within each simulated dataset and a standardized-maximum construction for both individual and joint containment. Write $G_c=\log R_c(4)$ and let $s_c$ be the sample standard deviation across bootstrap draws. The higher empirical 0.95 quantile of $\max_c |(G^*_{bc}-\widehat G_c)/s_c|$ is $q$, giving intervals $\widehat G_c\pm q s_c$. Joint containment requires all four intervals to contain their true configuration log ratios simultaneously. The full grid has 13 combinations of design, interval method, and target: percentile and BCa targets in the three P designs and the scalar monetary design, plus the four $\tau^2$-style members and their joint target.

\begin{table*}[t]
\centering
\caption{Synthetic interval coverage and delivery diagnostics.}
\label{tab:h4-calibration}
\begin{tabular}{llrrr}
\toprule
Design & Target & Coverage & MC 95\% & Simultaneous F13 \\
\midrule
Shared tie & percentile & 60/64 & [0.8476, 0.9827] & [0.7988, 0.9918] \\
Shared tie & BCa & 61/64 & [0.8691, 0.9902] & [0.8222, 0.9962] \\
Independent sham & percentile & 63/64 & [0.9160, 0.9996] & [0.8746, 1.0000] \\
Independent sham & BCa & 63/64 & [0.9160, 0.9996] & [0.8746, 1.0000] \\
High homogeneous & percentile & 61/64 & [0.8691, 0.9902] & [0.8222, 0.9962] \\
High homogeneous & BCa & 54/64 & [0.7314, 0.9224] & [0.6755, 0.9469] \\
$\tau^2$-style & c01 & 62/64 & [0.8916, 0.9962] & [0.8472, 0.9990] \\
$\tau^2$-style & c02 & 64/64 & [0.9440, 1.0000] & [0.9069, 1.0000] \\
$\tau^2$-style & c03 & 62/64 & [0.8916, 0.9962] & [0.8472, 0.9990] \\
$\tau^2$-style & c04 & 58/64 & [0.8070, 0.9648] & [0.7551, 0.9797] \\
$\tau^2$-style & joint & 54/64 & [0.7314, 0.9224] & [0.6755, 0.9469] \\
OpenHands-style & percentile & 60/64 & [0.8476, 0.9827] & [0.7988, 0.9918] \\
OpenHands-style & BCa & 62/64 & [0.8916, 0.9962] & [0.8472, 0.9990] \\
\bottomrule
\end{tabular}
\par\smallskip
\begin{minipage}{\linewidth}\footnotesize Coverage is the common raw and delivery count out of 64 independent simulated datasets for each combination of design, interval method, and target. Every delivered interval has positive width and all 13 combinations have zero delivery failures. Raw coverage allows finite ordered zero-width intervals containing truth; delivery also requires algorithm validity and positive width. Bounds are Clopper-Pearson Monte Carlo intervals for delivery only; F13 adjusts across 13 combinations. Methods and targets reuse the same datasets within each of the five designs, totaling 320 distinct simulated datasets, with no exclusions.\end{minipage}
\end{table*}

\subsection{Coverage and delivery}

\emph{Raw coverage} counts a finite ordered interval containing truth, including zero-width intervals. \emph{Delivery coverage} additionally requires an algorithm-valid interval of strictly positive width. Every planned simulated dataset stays in the denominator, including delivery failures. We report marginal 95 percent Clopper-Pearson Monte Carlo bounds for the 13 delivery probabilities, together with simultaneous bounds at level $1-0.05/13$, denoted F13. These bounds quantify the simulation's delivery probabilities rather than uncertainty in an agent effect. Raw coverage and delivery failures are counts only.

All 13 evaluated combinations have 64 independent simulated datasets, with no exclusions. Every interval is delivered successfully, so raw and delivery coverage counts coincide and no interval-delivery failures occur. Nevertheless, the homogeneous-control BCa target and the $\tau^2$-style joint target each have coverage of 54/64. Their marginal 95 percent Monte Carlo bounds are [0.7314, 0.9224], and their simultaneous F13 bounds are [0.6755, 0.9469]. Both shortfalls concern intervals missing truth, with their causes unresolved.

Table~\ref{tab:h4-calibration} reports all 13 evaluated combinations, including both interval methods wherever evaluated.

\subsection{Probability and cost laws}
\label{app:synthetic-laws}

The two task states have equal probability in every design. In P-\allowbreak SHARED-\allowbreak TIE and P-\allowbreak INDEPENDENT-\allowbreak SHAM, one state has success probability $9/10$ for every configuration; the other has probabilities, in fixed configuration order,
\begin{equation*}
\begin{aligned}
(&1/30,1/15,1/10,2/15,1/6,1/5,\\
 &1/5,4/15,3/10,1/3,11/30,2/5).
\end{aligned}
\end{equation*}
P-\allowbreak HIGH-\allowbreak HOMOGENEOUS sets both probabilities to $9/10$. All P designs use unit cost. The monetary designs use
\begin{equation*}
C=\operatorname{base}(Z,\mathrm{domain})(1+S)U,
\end{equation*}
with independent $U$ equally likely to be $1/2$ or $3/2$ on each attempt. M-\allowbreak TAU-\allowbreak HIGH uses success-probability pairs $(3/5,19/20)$, $(7/10,39/40)$, $(4/5,49/50)$, and $(9/10,99/100)$. Its state-specific base costs are $(4,1)$ on 50 airline tasks and $(6,2)$ on 114 telecom tasks. M-\allowbreak OPENHANDS-\allowbreak HIGH uses the last probability pair and base costs $(4,1)$.

\FloatBarrier
\balance
\bibliographystyle{ACM-Reference-Format}
\bibliography{references}

\end{document}